\documentclass[letterpaper, 10 pt, conference]{ieeeconf}  

\IEEEoverridecommandlockouts                              

\usepackage{graphics} 
\usepackage{epsfig} 
\usepackage{amsmath} 
\usepackage{amssymb}  
\usepackage{bm}
\usepackage{color}
\usepackage{mathtools}
\usepackage{ushort}
\usepackage{algorithm}
\usepackage[noend]{algpseudocode}
\usepackage{hyperref}

\usepackage{cite}

\title{\LARGE \bf
Safe planning and control under uncertainty for self-driving
}

\author{Shivesh Khaitan$^{1}$, Qin Lin$^{2*}$ and John M. Dolan$^{2}$
\thanks{$^{1}$Shivesh Khaitan is with the Department of Computer Science \& Engineering, Manipal Institute of Technology, Manipal, Karnataka, India
        {\tt\small shivesh.khaitan@learner.manipal.edu}}%
\thanks{$^{2}$Qin Lin and John M. Dolan are with the Robotics Institute, Carnegie Mellon University, Pittsburgh, PA, the USA
        {\tt\small \{qinlin,jdolan\}@andrew.cmu.edu}}%
\thanks{$^{*}$Corresponding author}
}

\begin{document}

\maketitle
\thispagestyle{empty}
\pagestyle{empty}

\begin{abstract}
Motion Planning under uncertainty is critical for safe self-driving. In this paper, we propose a unified obstacle avoidance framework that deals with 1) uncertainty in ego-vehicle motion; and 2) prediction uncertainty of dynamic obstacles from environment. A two-stage traffic participant trajectory predictor comprising short-term and long-term prediction is used in the planning layer to generate safe but not over-conservative trajectories for the ego vehicle. The prediction module cooperates well with existing planning approaches. Our work showcases its effectiveness in a Frenet frame planner. A robust controller using tube MPC guarantees safe execution of the trajectory in the presence of state noise and dynamic model uncertainty. A Gaussian process regression model is used for online identification of the uncertainty's bound. We demonstrate effectiveness, safety, and real-time performance of our framework in the CARLA simulator.
\end{abstract}

\section{INTRODUCTION}
Self-driving cars are expected to eliminate the possibility of accidents. This requires the algorithms used in self-driving to be robust against disturbances in a nondeterministic environment. In recent decades, motion planning research has considered uncertainty from specific sources. Still, there does not exist a framework that comprehensively assures safety across prediction, planning, and control layers in such environments.

The first research problem arises from the highly uncertain trajectories of surrounding vehicles due to sensing, localization, maneuver or intention uncertainties, etc. Our previous work proposed a two-stage prediction model \cite{pan2020}: the cooperative short-term predictor uses non-linear reachability analysis to provide a safe enclosure of all possible trajectories of dynamic obstacles; while the efficiency-oriented long-term predictor takes a long-term goal into account and avoids over-conservative planning. We have demonstrated its effectiveness of trajectory optimization using constrained ILQR (CILQR) in a Cartesian frame. However, such a module can also cooperate with other existing planning algorithms. In this work, we adapt such a prediction module to a Frenet frame planner, which is widely used in state-of-the-art self-driving architecture \cite{Apollo}.

The second research problem arises from the uncertainty of ego vehicle including 1) error in the assumed dynamic model; 2) disturbance of control commands; 3) state noise.

Safety of the autonomous system cannot be guaranteed unless such noise is taken into consideration. Model Predictive Control (MPC) \cite{2mpc, arab2016motion, mmpc,borrelli2005mpc} is one of the most successful approaches for system controls. The basic idea of MPC is to repeatedly solve optimization problems on-line to find an optimal input to the controlled system. Several MPC variants have been studied in recent decades which are robust to uncertainty. Overall, the robust algorithms can be divided into two classes of approaches: min-max techniques and robust tube MPC approaches. 

Min-max techniques optimize the worst-case performance of the controller with respect to bounded uncertainties. For a comprehensive literature review on min-max MPC techniques, readers are referred to \cite{lofberg2003minimax}. A general drawback of this technique is its worst case consideration. This exponentially increases the computational requirements, which is not suitable for real-time implementation.

The robust tube-based MPC techniques guarantee to keep the actual state within an invariant tube around the nominal MPC trajectory. They work by complementing the nominal MPC with a feedback controller which tries to keep the actual state within the tube by tracking the nominal trajectory. Tube-based approaches have been extensively studied in the literature \cite{mayne2005robust, rathai2017robust, jeantube, lopez2019dynamic, Mayne_2007}. However, these do not consider the presence of obstacles, in which non-convex state constraints need to be satisfied. \cite{Garimella_2018} and \cite{8796049} have proposed techniques for collision avoidance, but \cite{Garimella_2018} deals with static obstacles only, and \cite{8796049} uses invariant set computation for obstacles, which incurs high computation costs.



In this paper, we propose a computationally efficient unified safe planning and control framework for a non-linear system with the ability to avoid moving obstacles. We showcase the framework by adopting and improving the robust tube controller proposed in \cite{mayne2005robust} to guarantee control-level safety in three aspects: 1) The moving obstacles can be considered static across the planning horizon in a control cycle since the total simulation time is minimal; 2) We conduct convexification of the space around the ego vehicle by using IRIS (Iterative Regional Inflation by Semidefinite programming) \cite{Deits2015}, which can provide an obstacle-free convex region for efficiently solving an optimization problem; 3) Instead of setting the uncertainty bounds of tube MPC as user-defined parameters, which is unrealistic to obtain a priori, we use Gaussian Process Regression (GPR) for online identification.

\begin{figure}[thpb]
  \centering
  \includegraphics[width=\columnwidth]{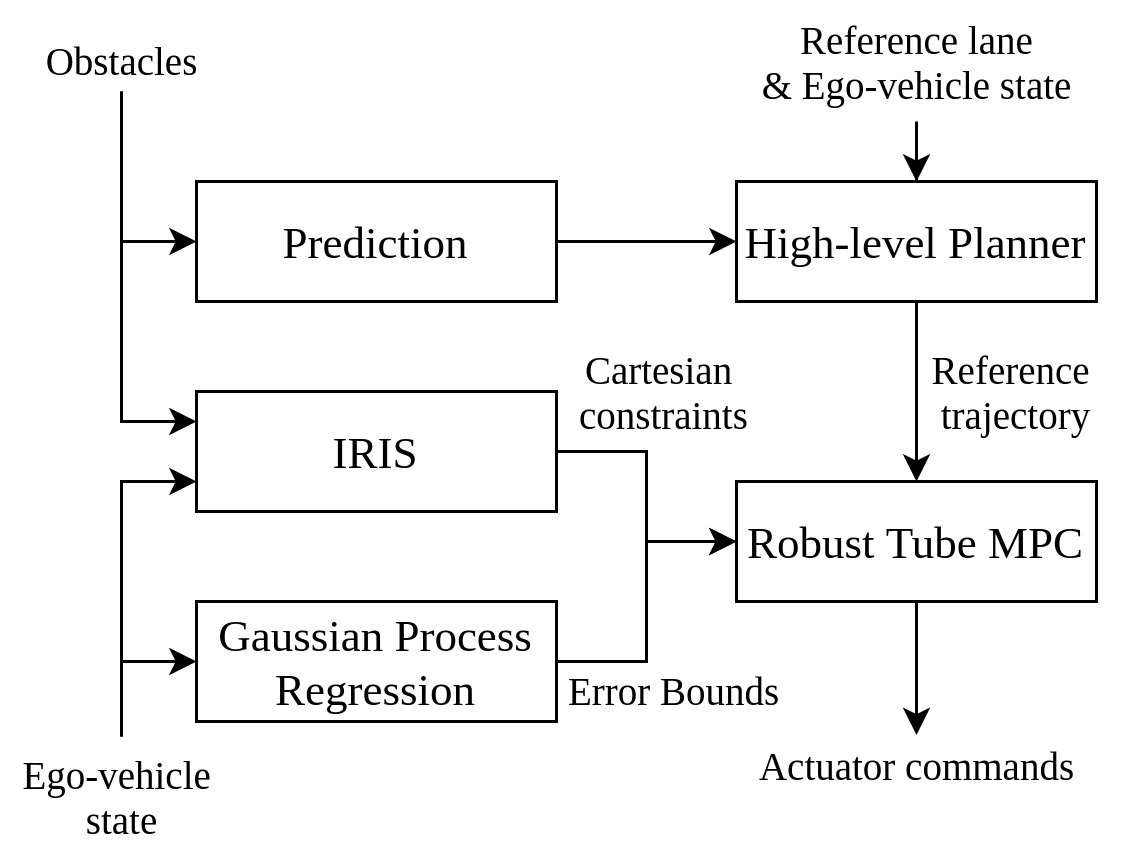}
  \caption{Overview of the framework}
  \label{fig:architecture}
\end{figure}

Fig. \ref{fig:architecture} shows a high-level overview of the proposed framework. Each obstacle's state, obtained from perception, is processed by the prediction module to predict its future states. The high-level planner then uses these predictions to generate a collision-free trajectory in the Frenet frame (hereafter referred to as ``frenet trajectory"). IRIS computes a convex obstacle-free region (hereafter referred to as ``IRIS region") around the ego vehicle, containing the reference points or choosing a closest point in the IRIS region from the upstream planned trajectory. The error bounds corresponding to the ego-vehicle current state are computed by the GPR module. The GPR model is trained using historic trajectories of the ego-vehicle. A linear time-variant (LTV) robust tube MPC uses the IRIS region as the original free state space and disturbance identified from GPR to compute control commands (acceleration and steering).

The rest of this paper is organized as follows. Section \ref{sec:related_work} provides a review of some important related work. Section \ref{sec:robust_planning_framework} describes the prediction and planning framework. Section \ref{sec:robust_control_framework} describes the low-level control framework. Section \ref{sec:experimental_results} presents the experimental results. The conclusions are in Section \ref{sec:conclusions}.

\section{RELATED WORK}
\label{sec:related_work}
Motion planning has been a widely studied research topic in the autonomous driving area for years. The main categories of approaches of motion planning include 1) graph search-based methods such as various shortest path algorithms and state lattice; 2) sampling-based methods such as rapidly-exploring random tree (RRT); 3) various curve interpolation-based methods; 4) optimization-based approaches using sequential programming \cite{gonzalez2015review}.

LQG-based approaches have shown good performance to deal with uncertainty of ego-vehicle state by combining Kalman filter and LQR. However, as dynamic obstacles, the movement uncertainty of surrounding traffic participants poses challenges to safe planning. Xu et al. proposed a framework to consider prediction uncertainty of dynamic obstacles and state and control uncertainty \cite{xu2014motion}. Safety performance was significantly improved with the general planning approaches without considering uncertainty. However, like most existing uncertainty-aware approaches, such an approach only has a probabilistic guarantee.

Reachability is a rigorous approach to provide a set enclosure of all feasible trajectories of a dynamic obstacle. Pek et al. \cite{pek2018computationally} used the set-based prediction tool SPOT \cite{koschi2017spot} to predict occupancy of all traffic participants and computed fail-safe trajectories using convex optimization. However, in our previous work \cite{pan2020}, we have shown that computing reachable sets of dynamic obstacles over the entire planning horizon leads to an over-conservative planner for general scenarios (i.e., cases other than an emergency requiring a fail-safe plan). Instead, we used a two-stage prediction consisting of the short-term and long-term. The short-term prediction leveraged reachability analysis while the long-term one is based on any data-driven predictor. Since the planning update time is smaller than the short-term horizon, safety can be guaranteed with a receding window. Meanwhile, viewing the long-term goal ahead significantly mitigates the over-conservative planning problem.

Unfortunately, collision-free planned trajectories do not necessarily guarantee that a vehicle will safely follow them due to disturbances in the control layer. Such disturbances can be from state noise or dynamic model error. Literature and engineering practice suggest including a margin (i.e., a buffer distance between planned trajectories and obstacles) in the feedback control \cite{werling2010optimal}. But the size of the margin has not been deeply investigated. Robust optimal control such as robust MPC has been applied to safe over-taking for autonomous driving \cite{dixit2019trajectory}. However, the scenario was over-simplified and the dynamic model's uncertainty was not considered.

In summary, the contribution of our work is that we propose a unified framework to consider: 1) the state and prediction uncertainty of dynamic obstacles in the prediction and planning layer; 2) state noise, control disturbance, and dynamic model uncertainty in the control layer.

\section{ROBUST PLANNING FRAMEWORK}
\label{sec:robust_planning_framework}
In this section, we will introduce the prediction and robust planning parts in the proposed framework. We use a short-term-and-long-term prediction to get the occupancy of dynamic obstacles at each time step for the planner to generate trajectories in free space.
\subsection{Dynamic Obstacles Prediction}
Prediction of moving obstacles is achieved using the short-term and long-term prediction model from \cite{pan2020}. It proposes a combination of a safety-oriented short-term planner and an efficiency-oriented long-term planner.

The short-term prediction considers the uncertainty of target-vehicle state (e.g., sensor disturbance or localization error) and the uncertainty of unobservable but bounded control actions over a short-term prediction horizon under a kinematically feasible but possibly non-deterministic assumption. The reachable state of the target vehicle is projected to the sub-space in the inertial frame as occupancy having the min-max longitudinal and lateral positions. The long-term predictor only predicts the target-vehicle position (i.e., particle) sequence without considering uncertainty. Using a pure particle-based prediction method over the whole planning horizon, like many existing machine learning-based predictors, safety is unfortunately not guaranteed. However, a pure set-based prediction method over the whole planning horizon will be over-conservative. The same horizon setting as in \cite{pan2020} is used in this work: for a planning horizon normally larger than 0.5 seconds, 0.5 seconds for the short-term, and the remaining for the long-term. The \emph{reachable set} of the system in a given time horizon $[0,T]$ is defined by $\underset{T}{\textbf{reach}}(\bm{x}(0), \mathcal{U}) \coloneqq \{\varphi_f(\bm{x}(0),\bm{u}(0), t)\,|\,\bm{x}_0\in \mathcal{X}_0, \bm{u}(t)\in \mathcal{U}, t\in [0,T]\}$, where $\varphi_f(\bm{x}_0, \bm{u}_0, t)$ is the solution of the ODE with the initial state $\bm{x}_0$ and under the control input $\bm{u}(t)$. However, for a general non-linear ODE, its explicit solution is non-trivial to obtain. An over-approximation using Taylor model is used to get the reachable set at every discrete time $\hat{\mathcal{R}}_{1}, \cdots, \hat{\mathcal{R}}_{k}$. The occupancy set of dynamic obstacle $i$ at time $k$ is represented by a box $\textbf{proj}(\hat{\mathcal{R}}^{i}_k) \coloneqq \textnormal{rectangle}(\ushort{p}_{x, k}^{i}, \bar{p}_{x, k}^{i}, \ushort{p}_{y, k}^{i}, \bar{y}_{y, k}^{i})$.

\subsection{Robust Frenet Frame Planning}
\begin{figure}[thpb]
  \centering
  \includegraphics[width=0.95\columnwidth]{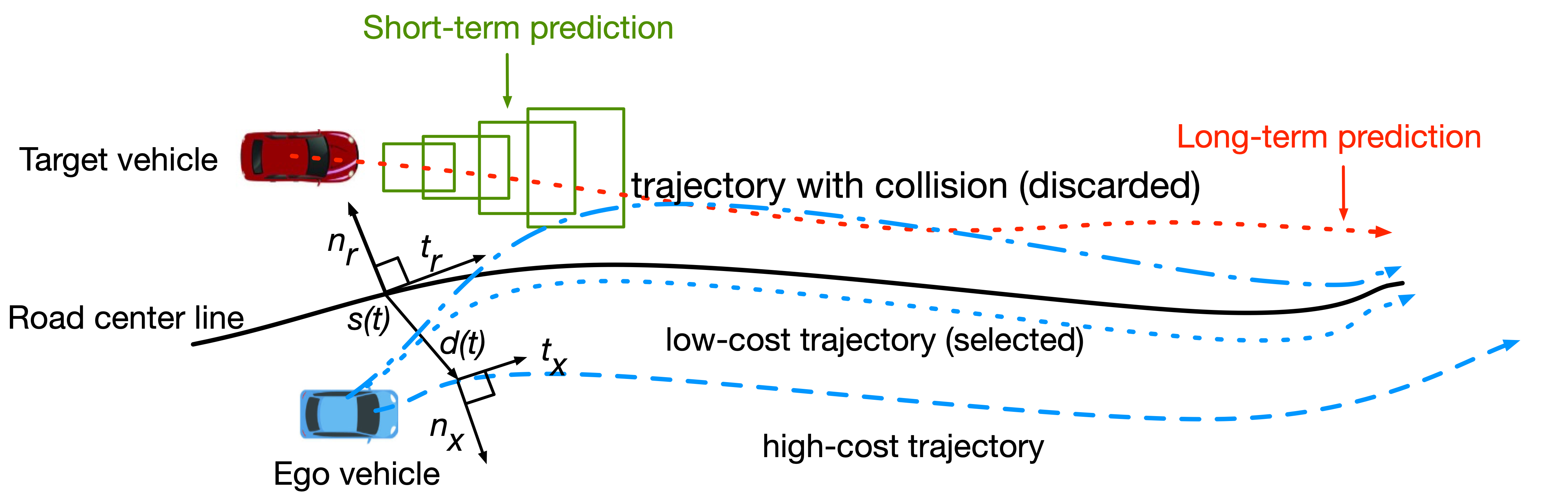}
  \caption{Collision-free trajectories planned in the Frenet frame. The blue example sampled trajectory at the top is discarded due to a detected collision. The middle blue trajectory has lower cost than the bottom one due to its closer distance to the reference.}
  \label{fig:frenet_planning}
\end{figure}

Frenet frame-based planning has been successful in practice due to the significant advantage of its independence from complex road geometry \cite{werling2010optimal}. As shown in Fig. \ref{fig:frenet_planning}, the lateral motion $d(t)$ and longitudinal motion $s(t)$ can be projected to the reference, which is usually the centerline of the road with arbitrary shape. The process of generating trajectories is essentially a curve interpolation method which combines a quintic polynomial for lateral motion and a quartic polynomial for longitudinal motion and generates an optimal collision-free trajectory by weighing the generated trajectories based on a combined cost-function. The final cost $(C)$ is as follows:
\begin{align}
    C_{lat} = k_j J_{lat} + k_a a_{lat} + k_v v_{lat} + k_s s_{lat} \nonumber \\
    C_{lon} = k_j J_{lon} + k_a a_{lon} + k_t T + k_s s_{lon} \nonumber \\
    C = k_{lat} C_{lat} + k_{lon} C_{lon} + k_{obs} s_{obs} \label{eqn:frenet_cost}
\end{align}
where $k_j, k_a, k_s, k_t, k_v, k_{lat}, k_{lon}, k_{obs} > 0$ are parameters, $J_{lat}$ and $J_{lon}$ are jerks, $a_{lat}$ and $a_{lon}$ are accelerations, $v_{lat}$ is the lateral velocity, $T$ is the time-taken, $s_{lat}$ and $s_{lon}$ are deviations from the reference and $s_{obs}$ is the sum of inverse distances from the nearest obstacle at each time step.

For dynamic obstacle avoidance, each of the trajectory points is checked for an overlap with the predicted obstacles at the trajectory's corresponding sampled time. For instance, the $(x, y)$ position of each trajectory at 0.1 seconds is checked for overlap with each of the predicted obstacles' positions at 0.1 seconds. The trajectories having overlapping points are discarded. The remaining trajectories are weighed according to Equation (\ref{eqn:frenet_cost}). Fig. \ref{fig:frenet_planning} shows an example of discarded, high-cost, and low-cost trajectories.

\section{ROBUST CONTROL FRAMEWORK}
In this section, we will introduce the robust control part of the proposed framework. We use a tube MPC to robustly follow the trajectory generated from the planner. The highly non-convex problem due to obstacles is solved by convexification. An online uncertainty identification using GPR will also be introduced.
\label{sec:robust_control_framework}
\subsection{Notation}
Given two sets $\mathbb{P}_1 \in \mathbb{R}^{n}$, $\mathbb{P}_2 \in \mathbb{R}^{m}$
\begin{itemize}
    \item $\mathbb{P}_1 \oplus \mathbb{P}_2$ represents the Minkowski set addition between the sets and $\mathbb{P}_1 \ominus \mathbb{P}_2$ represents the Pontryagin difference  between them.
    \item Let $\mathbb{I}$ be a subset of $\{1 \cdots n\}$. Then $\mathbb{P}_1 \downarrow \mathbb{I}$ represents $\mathbb{P}_1$'s projection on variables with indices in $\mathbb{I}$. 
    \item $\mathbb{P} = \mathbb{P}_1 \times \mathbb{P}_2$ represents the concatenation/conjunction of sets. Thus, $\mathbb{P} \in \mathbb{R}^{m+n}$.
\end{itemize}

\subsection{Problem Formulation}
\label{sec:problem_formulation}
A general optimal control problem subject to constants is formulated as: 
\begin{align}
& \underset{\bm{U}}{\text{min}}
& J = \sum_{k=0}^{N-1}l(\bm{x}_k, \bm{u}_k)+l_f(\bm{x}_N)  \label{eqn:objective_function}\\
& \text{s.t.}
& \bm{x}_{k+1} = A\bm{x}_k + B\bm{u}_k + \bm{w_k} \label{eqn:system_dynamics}
\\
& & \bm{x} \in \mathbb{X}\\
& & \bm{u} \in \mathbb{U} 
\end{align}

\noindent where $A \in \mathbb{R}^{m \times m}$ and $B \in \mathbb{R}^{m \times n}$ are dynamics and input matrices respectively for the ego vehicle. $\bm{x}_k$, $\bm{u}_k$ and $\bm{w}_k$ are the state, control input and disturbance respectively at time step $k$ and $\bm{x}_N$ is the final state.

$\mathbb{U} \subset \mathbb{R}^m$ and $\mathbb{X} \subset \mathbb{R}^n$ are control limit and state constraints. For an obstacle avoidance problem, we need an extra configuration constraint as $\mathbb{X} \downarrow \mathbb{C} \cap \mathcal{O} = \emptyset$, where $\mathcal{O}$ represents the obstacles' occupancy set and $\mathbb{X}$ has been projected into a subspace such as the Cartesian frame. The disturbance $\bm{w}$ is assumed to be bounded by $\bm{w}  \in \mathcal{W}$. Equation (\ref{eqn:system_dynamics}) is the system dynamics constraint, which is a transition function mapping state, control and disturbance at step $k$ to state at step $k+1$. $\bm{X}^{*} \coloneqq \{ \bm{x}_0, \bm{x}_1, \cdots, \bm{x}_{N} \}$ is the optimal state sequence. $l$ and $l_f$ are the cost functions.

\subsection{System Dynamics}
The model used in this work is a kinematic bicycle model. However, it can be replaced by a more complex dynamic model. The control input is $\bm{u}_k = [a_k, \delta_k]^T$,  where $a_k$ and $\delta_k$ are the acceleration and steering angle, respectively. Equation (\ref{eqn:dyneqn}) represents the transition function with non-zero steering angle. The horizontal increment is calculated by $\int_{0}^{l}\cos(\theta_k + \kappa s)ds$ and the vertical increment is calculated by $\int_{0}^{l}\sin(\theta_k + \kappa s)ds$. $\kappa = \frac{\tan(\delta)}{L}$ is the curvature, $L$ is the vehicle length, and $l_k = v_k \Delta t + \frac{1}{2}a \Delta t^2$ is the distance travelled at time $k$ with discretization step $\Delta t$.
\begin{align}
    p_{x, k+1}  &= p_{x, k} + \frac{\sin(\theta_{k} + \kappa l_k) - \sin{\theta_k}}{\kappa} \nonumber \\
    p_{y, k+1}  &= p_{y, k} + \frac{\cos{\theta_k} - \cos(\theta_{k} + \kappa l_k)}{\kappa} \nonumber \\
    v_{k+1}     &= v_{k} + a \Delta t \nonumber \\
    \theta_{k+1}&= \theta_{k} + \kappa l_k \label{eqn:dyneqn}
\end{align}
However, when $\delta$ is zero, $\kappa$ will be zero, which will cause a numerical issue in Equation (\ref{eqn:dyneqn}). Hence, a different update shown in Equation (\ref{eqn:dyneqn_zero}) is used. 
\begin{align}
    p_{x, k+1}  &= p_{x, k} + (v_t dt + \frac{1}{2} a \Delta t^2) \cos(\theta_k) \nonumber \\
    p_{y, k+1}  &= p_{y, k} + (v_t dt + \frac{1}{2} a \Delta t^2) \sin(\theta_k) \nonumber \\
    v_{k+1}     &= v_{k} + a \Delta t \nonumber \\
    \theta_{k+1}&= \theta_{k} \label{eqn:dyneqn_zero}
\end{align}


\subsection{Robust Tube MPC}
\label{sec:robust_tube_mpc}


Let $\bm{\bar{x}}$ and $\bm{\bar{u}}$ be the new state and input of a nominal system ignoring $\bm{w}$. Let ${K} \in \mathbb{R}^{m \times n}$ be such that $A_K = A + BK$ is stable. Let $\mathbb{Z}$ be a disturbance invariant set for the controlled uncertain system $\bm{x}^{+} = A_K\bm{x} + \bm{w}$, therefore satisfying $A_K \mathbb{Z} \oplus \mathcal{W} \subseteq \mathbb{Z}$. Then, robust tube MPC solves the following optimization problem for $\bm{\bar{X}}$ and $\bm{\bar{U}}$:
\begin{align}
& \underset{\bm{\bar{U}}}{\text{min}}
& \bar{J} = \sum_{k=0}^{N-1}l(\bm{\bar{x}}_k, \bm{\bar{u}}_k)+l_f(\bm{\bar{x}}_N) \\
& \text{s.t.}
& \bm{\bar{x}}_{k+1} = A\bm{\bar{x}}_k + B\bm{\bar{u}}_k \label{eqn:LTIV}\\
& & \bm{\bar{x}} \in \mathbb{X} \ominus \mathbb{Z} \label{eqn:x_z_bound}\\
& & \bm{\bar{u}} \in \mathbb{U} \ominus K\mathbb{Z} \label{eqn:u_kz_bound}\\
& & \bm{x_0} \in \bm{\bar{x}_0} \label{eqn:x0_bound} \oplus {\mathbb{Z}}
\end{align}

After solving the above, the final control is given by $\bm{u} = \bm{\bar{u}} + K(\bm{x} - \bm{\bar{x}})$, which guarantees $\bm{x}^{+} \in \bm{\bar{x}^{+}} \oplus \mathbb{Z}$ for all $\bm{w} \in \mathcal{W}$ where $\bm{x^{+}} = A\bm{x} + B\bm{u} + \bm{w}$. That is, it guarantees that the actual state $\bm{x_k}$ will always be inside the convex constraint set $\mathbb{X}$. Note that the formulation also applies to a LTV system, where $A$ and $B$ are replaced by Jacobian matrices $A_k$ and $B_k$ in Equation (\ref{eqn:LTIV}). For a detailed proof and study of feasibility and stability of the above controller, readers are referred to \cite{mayne2005robust}. The various constraints for the MPC are computed as follows: 

\subsubsection{State constraint ($\mathbb{X}$)}
The free space of velocity and yaw is readily computed by performing a difference of the original space and disturbance. However, it is not trivial for state variables $p_x$ and $p_y$. The highly non-convex space in the presence of obstacles is challenging for the control optimization. In this work, the cartesian constraint $\mathbb{X}_c$ for robust tube MPC is computed using IRIS \cite{Deits2015}, which gives a convex obstacle-free region containing the ego vehicle and the initial reference points of the frenet trajectory. IRIS alternates between two convex optimizations: 1) a quadratic program that generates a set of hyperplanes to separate a convex region of space from the set of obstacles; 2) a semidefinite program that finds a maximum-volume ellipsoid inside the polytope intersection of the obstacle-free half-spaces defined by those hyperplanes. Given the position of the ego vehicle as an initial seed point in space, around which the first ellipsoid is constructed, IRIS grows the ellipsoid greedily at every iteration until it reaches a local fixed point. The initial reference points from the frenet trajectory are set as required contained points. The final set of separating hyperplanes forms a convex polytope, which is a convex region of obstacle-free space. Finally, the full state constraint is computed as:
\begin{align}
    \mathbb{X} = \mathbb{X}_c \times \mathbb{V} \times \mathbb{Y}
\end{align}
where $\mathbb{V}$ and $\mathbb{Y}$ are velocity and yaw constraints respectively. They can be obtained a priori from the vehicle's configuration constraints.

\subsubsection{Control constraint ($\mathbb{U}$)}
Control constraints such as acceleration and steering are obtained from a vehicle's control limits.

\subsubsection{Disturbance invariant set ($\mathbb{Z}$)}
The disturbance invariant set is over-approximated using $\mathbb{Z} = \sum_{i=0}^{n}A_k^{i}\mathcal{W}$ as described in \cite{rakovic2004invariant}. 

\subsection{Efficient polytope operation}
In the tube MPC part, all sets are represented as polytopes. The manipulation of convex polytopes relies on the so-called \emph{dual description}. Depending on different operations, the vertex representation (V-rep) and the half-space representation (H-rep) can have significantly distinct efficiency. For example, a union of two polytopes is easily done by using the union set of all vertices; while the Minkowski sum and Pontryagin difference are easier for a V-rep. Unfortunately, a well-known problem is that with the growth of dimension, the dual transformation will grow exponentially.

For example, the resultant $\mathbb{Z}$ in V-rep $\mathbb{Z}_v$ needs be converted to H-rep $\mathbb{Z}_h$. This conversion for a 4-D polytope is computationally expensive. To deal with this, an over-approximated H-rep is calculated by projecting the vertex representation of $\mathbb{Z}$ to two independent planes $\mathbb{C}$ with $(p_x, p_y)$ variables and $\mathbb{C'}$ with $(v, \theta)$ variables, computing the H-rep separately, and concatenating them thereafter. Algorithm \ref{alg:invariant_set_half_space_representation} describes the proposed H-rep computation algorithm.

\begin{algorithm}
\caption{Fast H-rep transformation from V-rep for high-dimensional space}\label{alg:invariant_set_half_space_representation}
\begin{algorithmic}[1]
\Procedure{computeZ}{$A_k$, $\mathcal{W}$, n}
\State $\mathbb{Z}_v \gets \sum_{i=0}^{n}A_k^{i}\mathcal{W}$ \Comment{compute V-rep of $\mathbb{Z}$}
\State $\mathbb{Z}_v^{c} \gets \mathbb{Z}_c \downarrow \mathbb{C}$
\State $\mathbb{Z}_v^{c'} \gets \mathbb{Z}_c' \downarrow \mathbb{C'}$
\State $\mathbb{Z}_h^{c} \gets \textbf{computeH}(\mathbb{Z}_v^{c})$
\Comment{H-rep of $\mathbb{Z}_v^{c}$}
\State $\mathbb{Z}_h^{c'} \gets \textbf{computeH}(\mathbb{Z}_v^{c'})$
\Comment{H-rep of $\mathbb{Z}_v^{c'}$}
\State $\mathbb{Z}_h \gets \mathbb{Z}_h^{c} \times \mathbb{Z}_h^{c'}$
\Comment{Concatenate $\mathbb{Z}_h^{c}$ and $\mathbb{Z}_h^{c'}$}
\State \textbf{return} $\mathbb{Z}_h$
\EndProcedure
\end{algorithmic}
\end{algorithm}


\subsection{Estimation of Error Bounds ($\mathcal{W}$)}
A significant difficulty in applying tube MPC in practice is that it is unrealistic to expect to be able to obtain the uncertainty bounds a priori. Instead of setting the uncertainty bounds as user-defined parameters,  we estimate the disturbance $\bm{w}$ online using GPR \cite{7039601}. This is a non-parametric regression technique that extends multivariate Gaussian regression to the infinite-dimensional space of functions and provides a closed-form expression for Bayesian inference. It is a distribution over functions defined by a mean function $\mu\bm{(x)}$ and a covariance kernel function $k(\bm{x}, \bm{x}')$:
\begin{align}
    w(\bm{x}) \sim GP(\mu(\bm{x}), k(\bm{x}, \bm{x}'))
\end{align}

The posterior distribution of the function value $w(\bm{x})$ at a new point $\bm{x}_*$ can be calculated as:
\begin{align}
    w(\bm{x}_*) = \mu(\bm{x}_*) + KI^{-1} (Y - \mu(X)) \\
    var(\bm{x}_*) = K(\bm{x}_*,\bm{x}) - K(\bm{x}_*,X)KI^{-1}K(X,\bm{x}_*)\\
    KI = K(\bm{x}_*,X)(K(X,X) + \sigma_nI)
\end{align}
where $X = [\bm{x}_1 \cdots \bm{x}_n]^T$, $K_{ij}(X,X') = k(\bm{x}_i, \bm{x}'_j)$ and $Y = [\bm{y}_1 \cdots \bm{y}_n]^T$ are the labels for X. $\bm{x}$ is defined as $[v, \theta]$ since they are key variables in the Jaccobian matrix. $\bm{y}$ is the residual between the observed ego-vehicle state and the predicted state using the vehicle dynamic model. The GP has a zero mean function and a squared exponential covariance function. The error bounds $\mathcal{W}$ at the current state $\bm{x}_*$ are then calculated as:
\begin{align}
    \mathcal{W}(\bm{x}_*) = [w(\bm{x}_*) - c\sigma_*(\bm{x}), w(\bm{x}_*) + c\sigma_*(\bm{x})]
\end{align}
where $\sigma{(\bm{x}_*)} = \sqrt{var(\bm{x}_*)}$ is the standard deviation, $c$ is a parameter.

\section{EXPERIMENTAL RESULTS}
\label{sec:experimental_results}
In this section, we present our experimental results on the CARLA\cite{Dosovitskiy17} simulator. The vehicle used in the simulation is a Ford Mustang. We have set up experiments in multiple scenarios, including overtaking, intersections and curves. The disturbance injected into state and control action is bounded: $\bm{w} \in \mathcal{W} = \{ \bm{w}\; |\; \|\bm{w}\|\leq \bm{w}_{max}\}$ where $\bm{w}_{max} = 0.1$ for yaw and $\bm{w}_{max} = 0.2$ otherwise. Note that such a bound is unknown for the proposed algorithms and needs to be estimated. 

The ego-vehicle perceived position (with added disturbance) is pictured in purple with its transparency decreased over time. Its actual trajectory is plotted in blue. Target vehicles are depicted in red and the transparency works the same way. The IRIS regions are pictured as polygons in green. The planning horizon for the frenet trajectory is set to 3-4 seconds with a discretization of 0.1 seconds. For MPC, the planning horizon is 0.25 seconds with a discretization of 0.05 seconds.

We present the following three scenarios below:
\begin{enumerate}
    \item The ego vehicle is travelling behind a slow-moving target vehicle. It safely overtakes the target vehicle in the presence of other vehicles in the adjacent lanes.
    \item The ego vehicle is travelling through an uncontrolled intersection. It safely maneuvers around a target vehicle moving in a direction perpendicular to its motion.
    \item The ego vehicle overtakes a target vehicle on a curved road.
\end{enumerate}

The video demonstrations of the scenarios are available online\footnote{https://sites.google.com/view/safe-planning-and-control}. 
Following are the descriptions and results of the above scenarios:

\subsubsection{Overtake}
\begin{figure}[thpb]
  \centering
  \includegraphics[width=0.95\columnwidth]{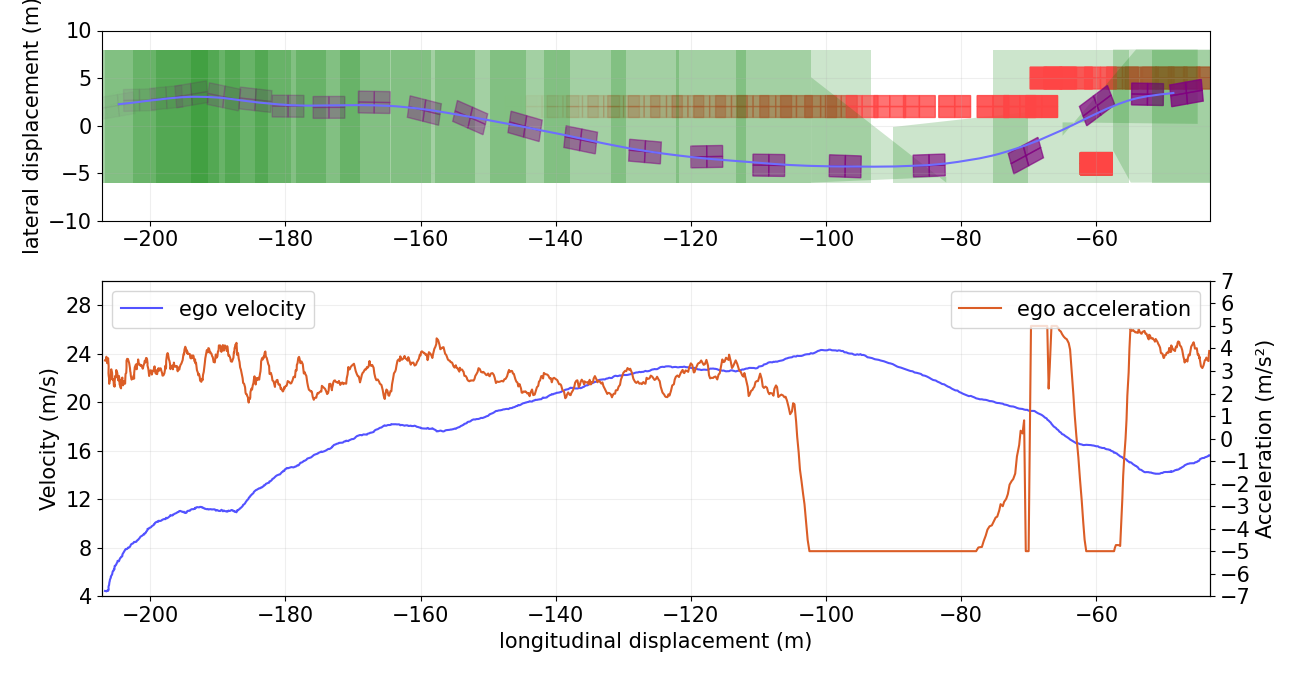}
  \caption{Overtaking in presence of obstacles in adjacent lanes}
  \label{fig:carla_overtake_uncertain}
\end{figure}

In this scenario, a target vehicle is travelling in front of the ego vehicle at 8m/s. The ego vehicle starts from rest and accelerates while approaching the target vehicle to achieve the target velocity of 25m/s. As seen in Fig.  \ref{fig:carla_overtake_uncertain}, the ego vehicle first decelerates to keep safe-distance from the target vehicle and begins the lane change. After passing the target vehicle it is faced by another parked vehicle. Without any noise, this is a comparatively easier situation in which the ego vehicle can just steer enough to move back to its reference lane. But due to high disturbance at this point, it deviates from the center of the lane. The robust control is, however, able to deal with the noise. The vehicle decelerates and safely steers back to the original orientation before beginning to accelerate again.

\subsubsection{Intersection}
\begin{figure}[thpb]
  \centering
  \includegraphics[width=0.95\columnwidth]{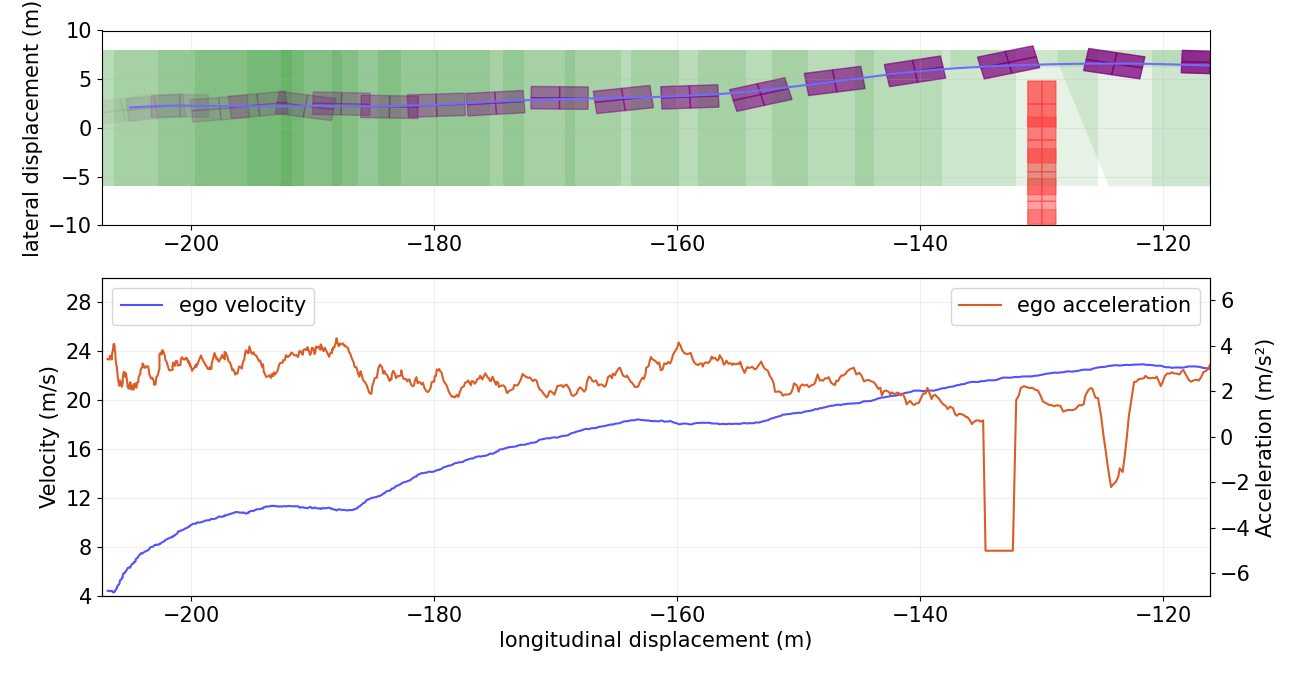}
  \caption{Dynamic obstacle avoidance in an intersection}
  \label{fig:carla_intersection_uncertain}
\end{figure}

In this scenario, a target vehicle is travelling perpendicular to the ego vehicle in an uncontrolled intersection at 5m/s. The ego vehicle starts from rest and is accelerating while approaching the intersection to achieve the reference velocity of 25m/s. As seen in Fig. \ref{fig:carla_intersection_uncertain}, at around a perpendicular distance of 5 meters from the target vehicle, the ego vehicle starts to steer to avoid the target vehicle. It drops the acceleration initially while turning, but starts increasing the acceleration again just as it nears the target vehicle. After avoiding the vehicle, it steers towards the reference lane. 

It can be noticed that there is a huge error in the perceived yaw of the ego vehicle when it nears the target vehicle. Even in this situation, since the error is bounded, the robust tube MPC is able to safely traverse the intersection.

\begin{figure}[thpb]
  \centering
  \includegraphics[width=0.95\columnwidth]{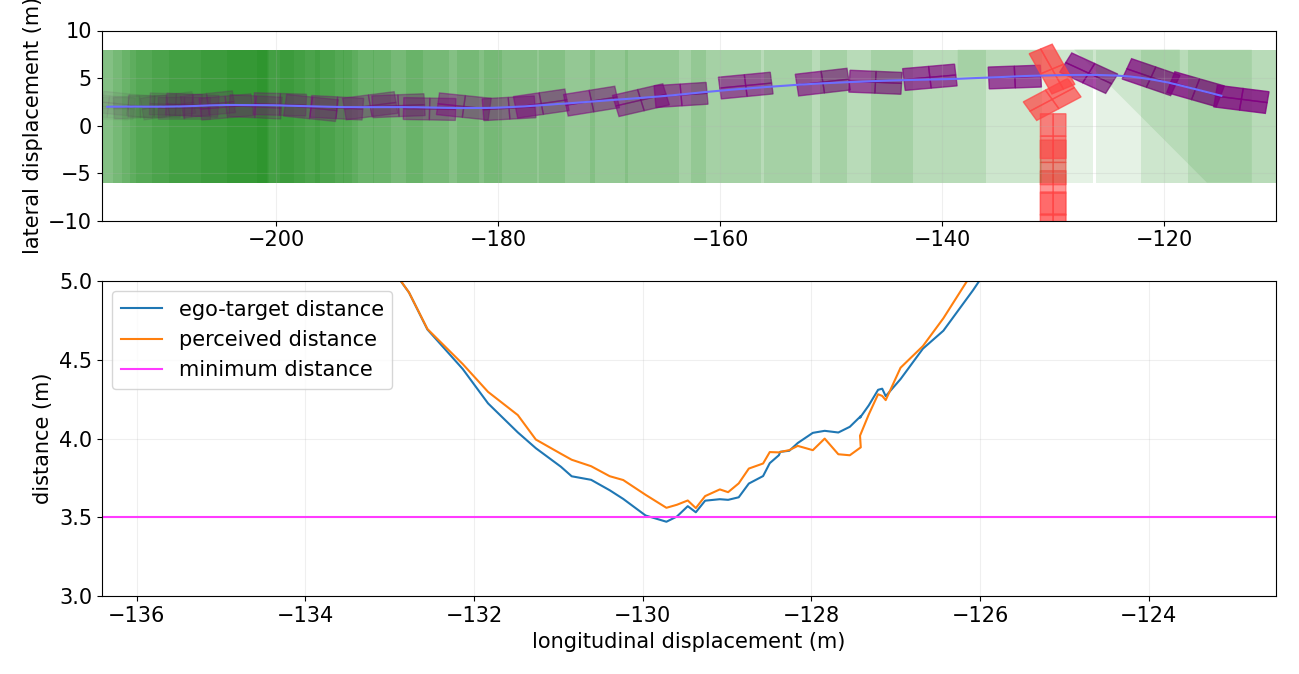}
  \caption{Collision at intersection when using nominal MPC. The ``zigzag" behavior of the target vehicle is due to the collison.}
  \label{fig:carla_intersection}
\end{figure}

In a similar scenario, the nominal MPC without considering uncertainty fails and collides with the target vehicle. As seen in Fig. \ref{fig:carla_intersection}, the nominal MPC drives the ego vehicle very close to the target vehicle and when the state noise is high, it is unable to avoid collision. The collision happens due to an error in state estimation which over-estimates the distance of the target vehicle from the ego vehicle. The second plot in Fig. \ref{fig:carla_intersection} shows the actual and perceived distances between the vehicles around the time of collision ($x = -130$ m). The minimum safe-distance between the ego vehicle and the target vehicle, determined based on the size of the vehicles, is 3.5 m. However, while moving past the target vehicle at the intersection, the actual distance (greater than the perceived distance due to error in state estimation) drops below the safe-distance and the vehicles collide.

\subsubsection{Curved road}
\begin{figure}[thpb]
  \centering
  \includegraphics[width=0.95\columnwidth]{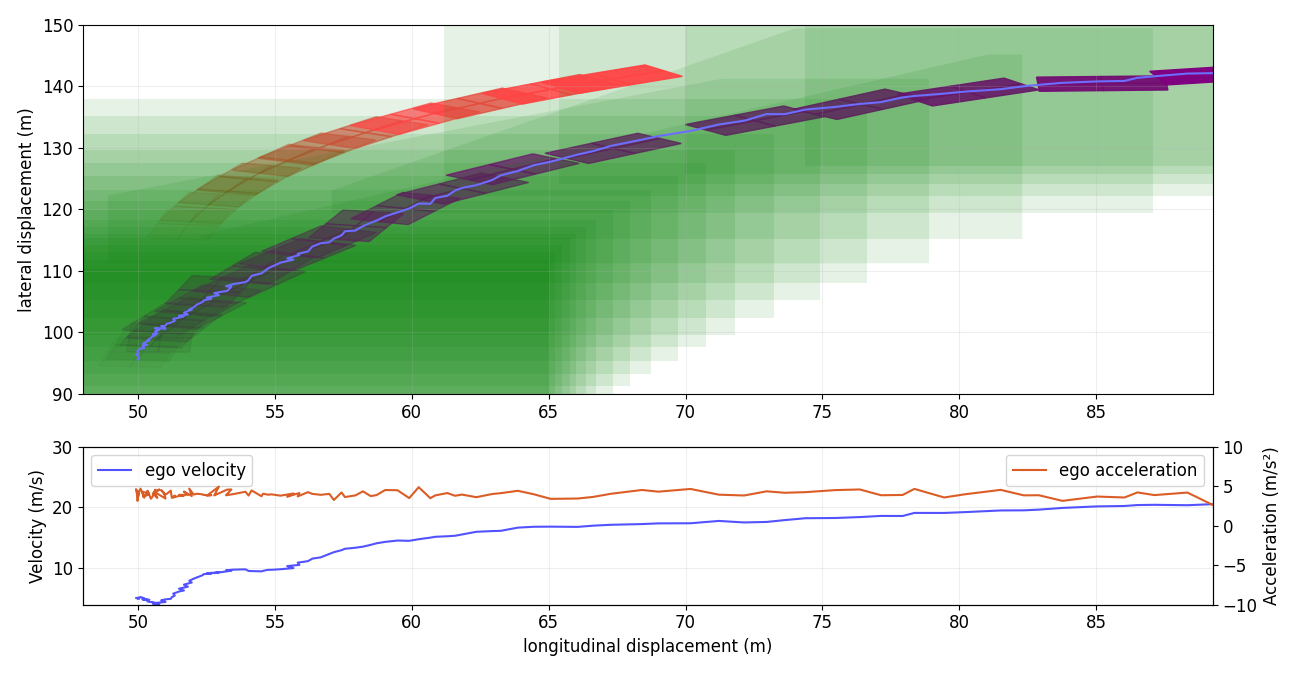}
  \caption{Overtaking in a curved road}
  \label{fig:carla_curve_uncertain}
\end{figure}

All scenarios above are tested on straight lanes. In this scenario, a slow vehicle is initially 10 meters ahead of the ego vehicle on a curved road. The target vehicle is travelling at 6 m/s and the ego vehicle successfully overtakes the target vehicle in the curve while accelerating to achieve the target velocity of 20 m/s. During the overtake, the ego vehicle is travelling at 15m/s and then it steers back to its reference lane and accelerates to achieve the reference velocity as seen in the velocity profile of Fig. \ref{fig:carla_curve_uncertain}. 

The average loop time for the MPC is 20ms, which makes it suitable for real-time implementation. The framework is implemented in C++. Gurobi \cite{gurobi} is used for solving the MPC. The control framework runs on a laptop with a 2.50GHz Intel Core i5-7200U CPU.

\section{CONCLUSIONS}
In this work, we propose a unified obstacle avoidance framework with safety assurance dealing with prediction uncertainty and ego vehicle uncertainty. Thanks to our robust prediction, the planner can efficiently generate collision-free trajectories. Using a tube MPC controller with significant improvements on efficient high-dimensional set operation and on-line uncertainty estimation, the vehicle is capable of robustly following the planned trajectories. Realistic scenarios have been tested in the high-fidelity CARLA simulator. The simulation results demonstrate the safety and real-time performance of our framework.
\label{sec:conclusions}


\addtolength{\textheight}{-12cm}   



\section*{ACKNOWLEDGMENT}

This material is based upon work supported by the United States Air Force and DARPA under Contract No. FA8750-18-C-0092. Any opinions, findings and conclusions or recommendations expressed in this material are those of the author(s) and do not necessarily reflect the views of the United States Air Force and DARPA.



\bibliographystyle{IEEEtran}
\bibliography{./IEEEfull,refs}

\end{document}